\documentclass{article}

\usepackage[margin=1in]{geometry}
\usepackage{graphicx}
\usepackage{amsmath}
\usepackage{amssymb}
\usepackage{caption}
\usepackage{subcaption}
\usepackage{tikz,pgfplots}
\usepackage{xcolor, colortbl}

\begin{document}

\title{Unsupervised Automated Event Detection using an Iterative Clustering based Segmentation Approach}

\author{Deepak K. Gupta\footnote{Email: GuptaDeepak2806@gmail.com}, Rohit K. Shrivastava, Suhas Phadke and Jeroen Goudswaard\\
Shell Technology Centre, Bengaluru, KA, India\\
}
\maketitle

\begin{abstract}
 A class of vision problems, less commonly studied, consists of detecting objects in imagery obtained from physics-based experiments. These objects can span in \mbox{4D $(x, y, z, t)$} and are visible as disturbances (caused due to physical phenomena) in the image with background distribution being approximately uniform. Such objects, occasionally referred to as `events', can be considered as high energy blobs in the image. Unlike the images analyzed in conventional vision problems, very limited features are associated with such events, and their shape, size and count can vary significantly. This poses a challenge on the use of pre-trained models obtained from supervised approaches.
 
In this paper, we propose an unsupervised approach involving iterative clustering based segmentation (ICS) which can detect target objects (events) in real-time. In this approach, a test image is analyzed over several cycles, and one event is identified per cycle. Each cycle consists of the following steps: (1) image segmentation using a modified $k$-means clustering method, (2) elimination of empty (with no events) segments based on statistical analysis of each segment, (3) merging segments that overlap (correspond to same event), and (4) selecting the strongest event. These four steps are repeated until all the events have been identified. The ICS approach consists of a few hyper-parameters that have been chosen based on statistical study performed over a set of test images. The applicability of ICS method is demonstrated on several 2D and 3D test examples.
	
\end{abstract}

\section{Introduction}

Machine learning can be used to detect `irregularities' in a provided dataset, and this process has been referred in the literature as anomaly detection \cite{Goldstein2016po}. In the context of real-time monitoring, an anomaly, referred to as `event', would be a disturbance observed in space and time for measurements done as a part of a certain experiment. Examples of automated event detection include identifying unusual human actions in a video \cite{Ke2007iccv}, detection of intrusion in internet traffic \cite{Sinclair1999} and automated picking of events in time-series seismic data \cite{Murat1992gp}. Several algorithms exist in the literature that can be used to detect such anomalies. For an overview of such methods, see the review works presented in \cite{Chandola2009acmcs, Goldstein2016po}.

\begin{figure}[t]
	\centering
	\begin{subfigure}{0.4\textwidth}
		\centering
		\includegraphics[scale=0.7]{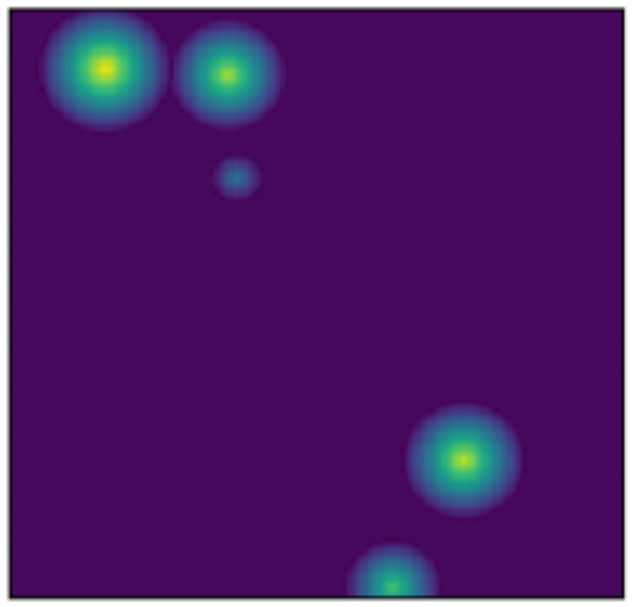}
		\caption{5 events}
	\end{subfigure}
	\begin{subfigure}{0.4\textwidth}
		\centering
		\includegraphics[scale=0.7]{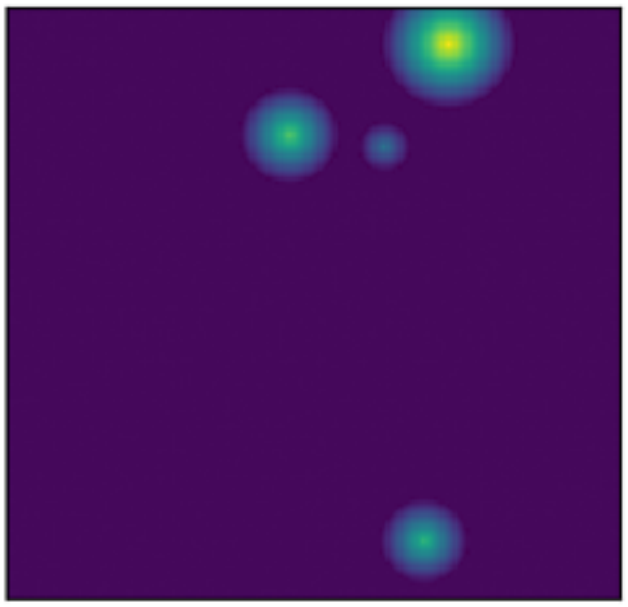}
		\caption{4 events}
	\end{subfigure}
	\caption{Schematic two-dimensional (2D) images containing (a) 5 energy events, and (b) 4 energy events. Each energy event is characterized by its source location in 2D space, origin amplitude, and spreading of the event.}
	\label{fig_syn}
\end{figure}

In event detection problems, the event itself can be considered as an anomaly compared to the rest of the data. For example, earthquakes are seismic events that show up as anomalous points on the seismogram \cite{Douglas1997gji}. These seismograms can be converted to 4D images ($x, y, z, t$) of the subsurface, and machine learning can provide real-time information of these earthquakes from the images. \mbox{Fig. \ref{fig_syn}} shows two examples of synthetic 2D images containing 5 and 4 events. The events are characterized as high amplitude zones, and the background has low amplitudes. The goal of this paper is to develop an automated approach to efficiently pick these events.

The problem of event detection as described above can be categorized under anomaly detection techniques dealing with images \cite{Augusteijn2002, Theiler2003}. Thus, we interchangeably use `anomaly' and `event' to refer to the class of problems described in this paper. The problem of event detection has been studied in the past using supervised as well as unsupervised methods \cite{Chandola2009acmcs}. The primary challenge with supervised learning approaches is to prepare the extensive training data that can make the trained model robust enough. Unsupervised methods partition the data into event and non-event classes. However, events in general are rare compared to the normal data, and due to this imbalance, the traditional unsupervised approaches suffer from high false alarm \mbox{rates \cite{Chandola2009acmcs}}.

Among the various unsupervised methods, clustering has widely been used to detect anomalous features in an image \cite{Scarth1995}. Without the need of any prior training, clustering methods group similar instances into groups, separating the non-event data from the events. For problems involving point or conceptual anomalies, clustering can easily identify the events \cite{Chandola2009acmcs}. However, for the type of problems considered in this paper, although a certain data point might not be an anomaly, a collection of data points lying together in ($x, y, z, t$) can constitute an event in the context of background values. Along this line, the traditional clustering algorithms can separate high amplitude points from the background (non-event) parts, however, the events themselves cannot be identified uniquely. Moreover, several other inherent drawbacks of these methods make them unsuited for the class of event detection problems dealt in this paper. 

We propose an iterative clustering based segmentation approach for the efficient identification of events in images. The unsupervised ICS approach decomposes an event detection problem into sub-problems: (1) image segmentation and (2) statistical interpretation of each segment. For segmentation purpose, clustering methodology is employed. We present a modified $k$-means clustering approach tailored to our image segmentation problem. Further, the amplitude distribution within each cluster is analyzed to decide whether a segment corresponds to an event or not. Details related to the proposed ICS methodology are discussed in Sections \ref{sec_desc} and \ref{sec_method}. The applicability of ICS method is demonstrated on 2D and 3D synthetic test images. Further, ICS is used to identify earthquake-events from a real 3D dataset. Related results are presented in Section \ref{sec_appl}, and a brief discussion on ICS method is presented in Section \ref{sec_discuss}. 

\begin{figure*}[t]
	\centering
		\includegraphics[scale=0.57]{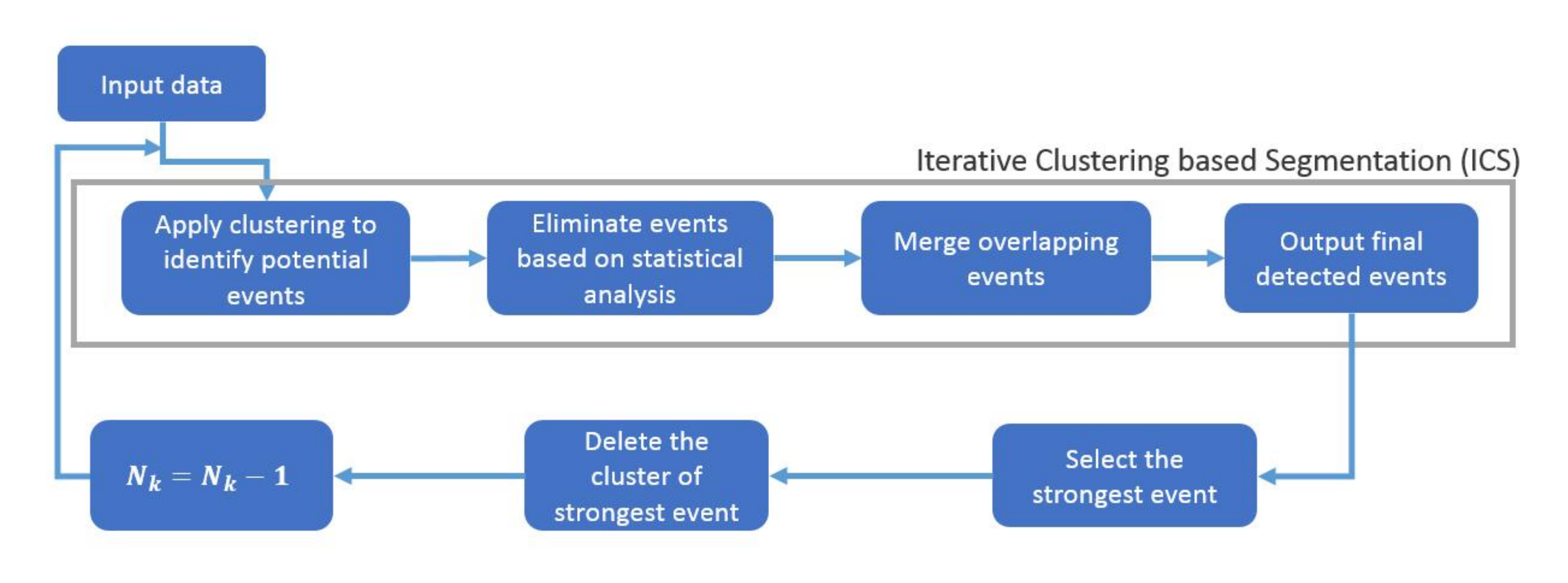}
	\caption{Flowchart for the detection of events using iterative clustering based segmentation (ICS) approach. Here, $N_k$ denotes the number of clusters used in the respective cycle. For clustering purpose, a variant of $k$-means clustering is used.}
	\label{fig_flow}
\end{figure*}
\begin{figure}[t]
	\centering
	\includegraphics[height=4.8cm, width=5.3cm]{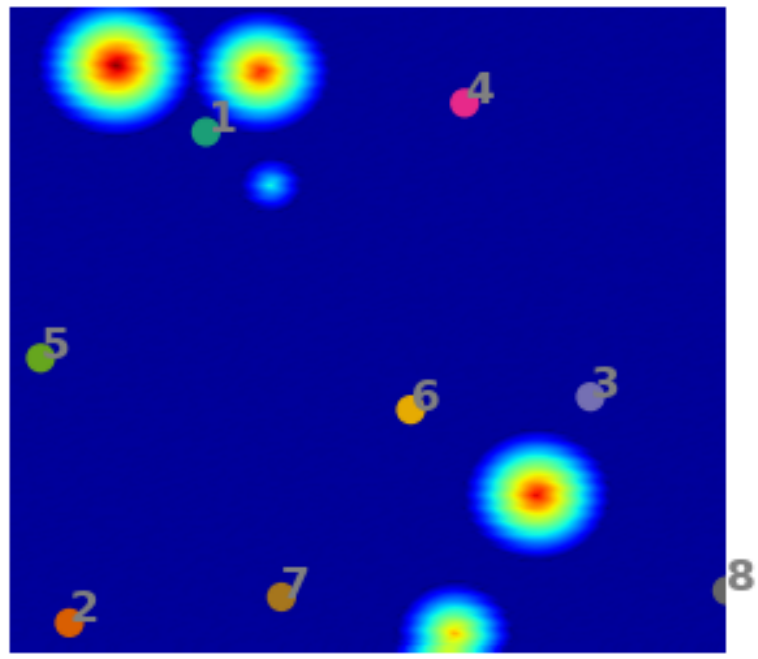}
	\includegraphics[height=4.8cm, width=5.3cm]{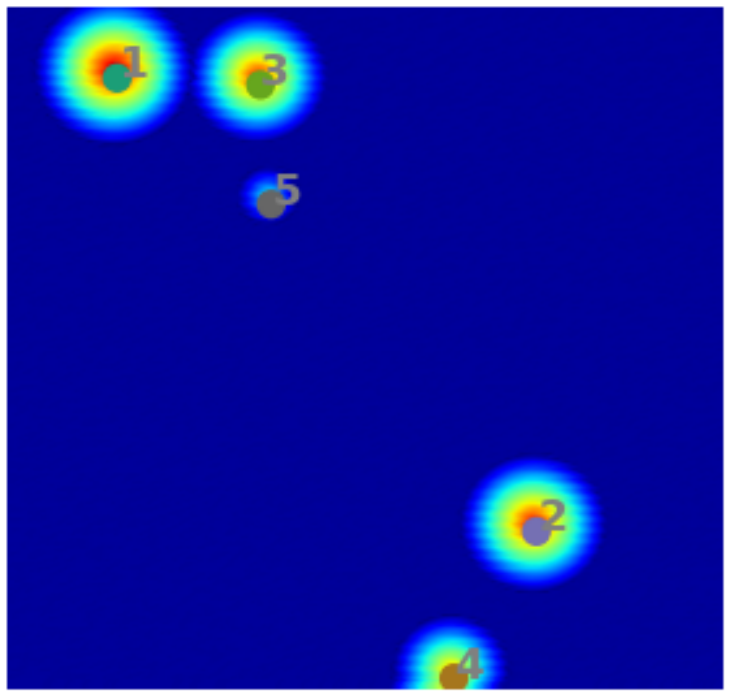}
	\caption{Initial centroids obtained using $k$-means++ (left) and their optimized locations (right) for a synthetic 2D image containing 5 events.}
	\label{fig_eg_kmeanspp}
\end{figure}
\section{Related Work}
Event (anomaly) detection problems, in general, have widely been studied in the field of computer vision. For an overview of the methods used to handle this problem, see the surveys presented in \cite{Goldstein2016po, Chandola2009acmcs}. These can primarily be divided into two categories: supervised methods and unsupervised methods. In the context of image processing, methods have been presented to identify events distributed in space and time., \emph{e.g.} satellite imagery \cite{Augusteijn2002}, spectroscopy \cite{Chen2005as} and mammographic image analysis \cite{Spence2001}. Some of the earliest methods used for automated event detection are mixture of models \cite{Spence2001}, regression methods \cite{Chen2005as}, Bayesian networks \cite{Diehl2002}, support vector machines \cite{Davy2002}, neural networks \cite{Augusteijn2002}, clustering \cite{Scarth1995} and nearest neighbor based techniques \cite{Pokrajac2007}.  Recently, several deep learning methods have also been proposed that can be used for detection (\emph{e.g.} \cite{Zhou2015, Cho2015, Siva2013}).

In general, the supervised algorithms can outperform the unsupervised algorithms in terms of speed, however, they are fragile to event detection problems \cite{Gornitz2013jair}. There exist several semi-supervised algorithms that can help to circumvent this issue, but, the unsupervised methods are still preferred due to their generalizability in identifying a wide-variety of events. Several unsupervised methods have been used in the past for event detection (\emph{e.g.} \cite{Chhabra2008, Aggarwal2013}). Based on the methodology employed, these can be categorized into divisions: nearest-neighbor based \cite{Ramaswamy2000}, clustering based \cite{He2003prl}, statistical \cite{Amer2013}, subspace based \cite{Kwitt2007} and classifier based \cite{Goldstein2012}. 

The nearest-neighbor algorithms such as $k$-NN approach perform well for classification problems, however, these are not very robust for an unsupervised setting \cite{Goldstein2016po}. Local variants such as local outlier factor (LOF) and local outlier probability (LoOP) approaches present a relative measure where the distance to $k$-nearest neighbors is compared to a reference value obtained for the entire data \cite{Breunig2000, Kriegel2009}. However, most of these algorithms suffer from the setback of choosing $k$ \cite{Goldstein2016po}. Cluster-based detection algorithms such as CBLOF \cite{He2003prl} and CMGOS use clustering to identify dense areas in the data, and in the next step, outliers are identified based on density estimates obtained for each cluster. However, for problems studied in this paper, data points are distributed almost uniformly, and the traditional clustering approach cannot be deployed to identify the events. 

Overall, to the best of our knowledge, the existing anomaly detection methods are not directly applicable to our event detection problems, and a tailored approach has been missing. In this work, we propose an iterative clustering based segmentation (ICS) approach to deal with these problems. 

\section{Iterative clustering based segmentation}
\label{sec_ics}
\subsection{Description}
\label{sec_desc}

The iterative clustering based segmentation (ICS) method analyzes the data over several cycles, and identifies one event per cycle. Fig. \ref{fig_flow} provides a schematic description of the ICS approach. During every cycle, ICS employs spatial clustering schemes with tailored assignment function, followed by statistical analysis of the data contained within each cluster. The statistical distribution within every cluster is studied and the cluster containing the strongest event is identified. The associated event is selected and the data related to this cluster is deleted. The whole process of event search is then repeated to find the next event. This process is continued till no more event can be identified in the remaining dataset. 

\begin{figure}[t]
	\centering
	\begin{subfigure}{0.4\textwidth}
		\centering
		\includegraphics[height=4.8cm, width=5.3cm]{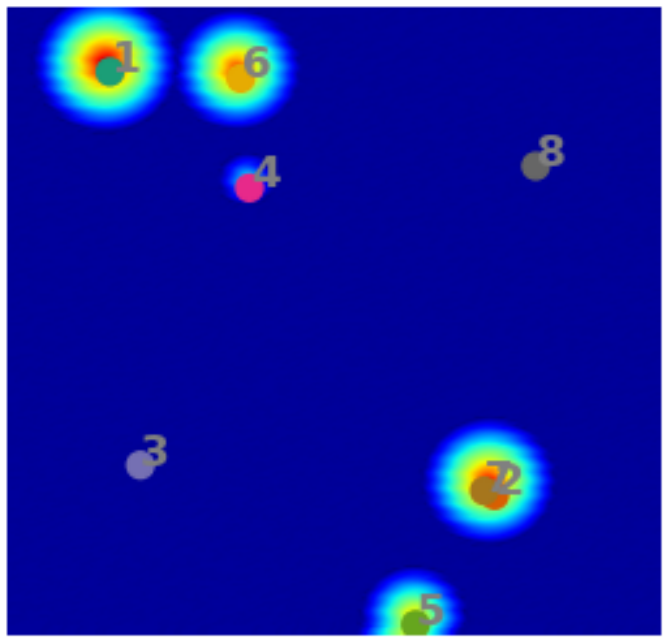}
		\caption{Identified events}
	\end{subfigure}
	\begin{subfigure}{0.4\textwidth}
		\centering
		\includegraphics[height=4.8cm, width=5.3cm]{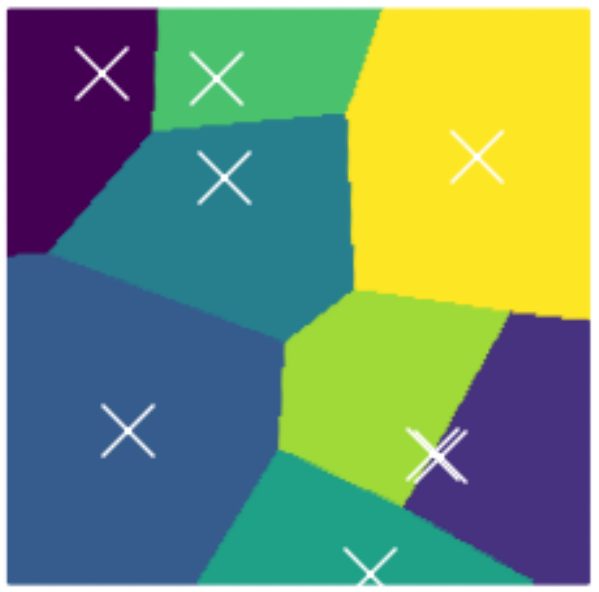}
		\caption{Optimized clusters}
	\end{subfigure}			
	\caption{Results obtained from modified $k$-means clustering during the first cycle of ICS, showing (a) the 5 synthetic events and the optimized locations of 8 cluster centers, and (b) the 8 segments with their centers marked using `$\times$'.}
	\label{fig_case1_kmeans}
\end{figure}

\begin{figure}[t]
	\centering
	\begin{subfigure}{0.4\textwidth}
		\centering
		\includegraphics[height=4.8cm, width=5.3cm]{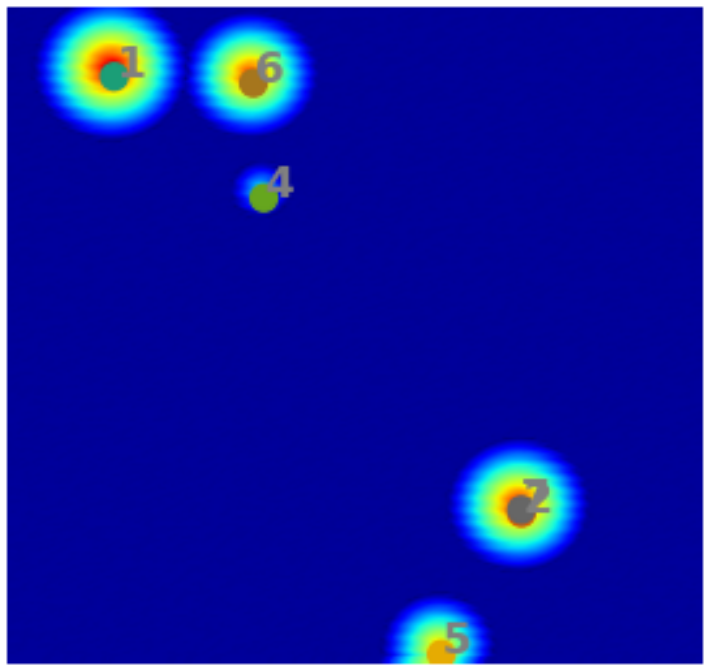}
		\caption{Identified events}
	\end{subfigure}
	\begin{subfigure}{0.4\textwidth}
		\centering
		\includegraphics[height=4.8cm, width=5.3cm]{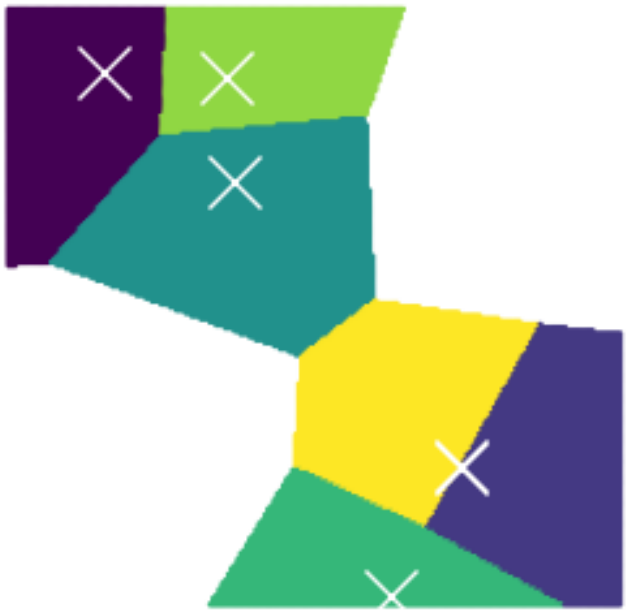}
		\caption{Optimized clusters}
	\end{subfigure}			
	\caption{Results obtained after the elimination of empty clusters in Fig. \ref{fig_case1_kmeans}, showing (a) a synthetic 2D image with 5 events overlapping with the remaining cluster-centers and (b) the remaining partitions. }
	\label{fig_kmeans_elim}
\end{figure}

For the purpose of clustering, a variant of the traditional $k$-means clustering has been used \cite{MacQueen1967}, and cluster centroids are initialized using the $k$-means++ \mbox{algorithm \cite{Arthur2007}}. \mbox{Fig. \ref{fig_eg_kmeanspp}} shows an example of a synthetic 2D test case, where the initial cluster centroids (obtained using $k$-means++) as well as their optimized locations are shown. The hyper-dimensions for clustering comprise spatial (as well as temporal dimensions in 4D) and the amplitude feature weighted by a power term $p$. Including the effect of amplitudes in clustering forces the centroids to lie close to centers of the events in the optimized segmentation, and ths impact is controlled by $p$. An example of event-centers and the associated partitions for a 2D problem is shown in Fig. \ref{fig_case1_kmeans}. More details on this aspect as well as the mathematical formulation are presented in \mbox{Section \ref{sec_method}}.

\begin{figure*}[t]
	\centering
	\begin{subfigure}{0.47\textwidth}
		\includegraphics[scale=0.5, trim = 14.7cm 0 0 0, clip=true]{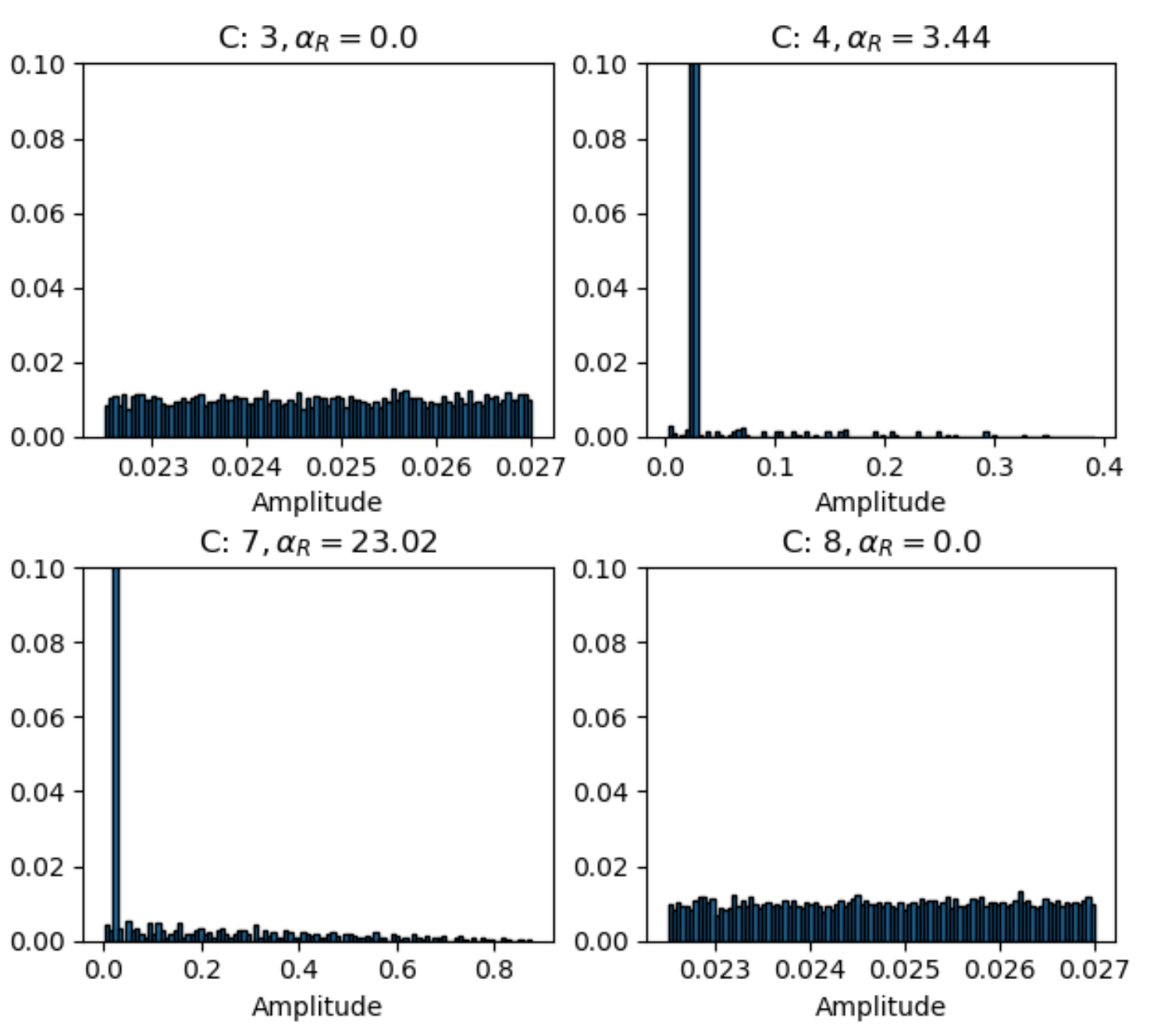}
		\caption{Noisefree data}
		\label{fig_stat1}
	\end{subfigure}
	\begin{subfigure}{0.47\textwidth}
	\includegraphics[height=6.6cm, width=7.6cm, trim = 0 0 14.97cm 0, clip=true]{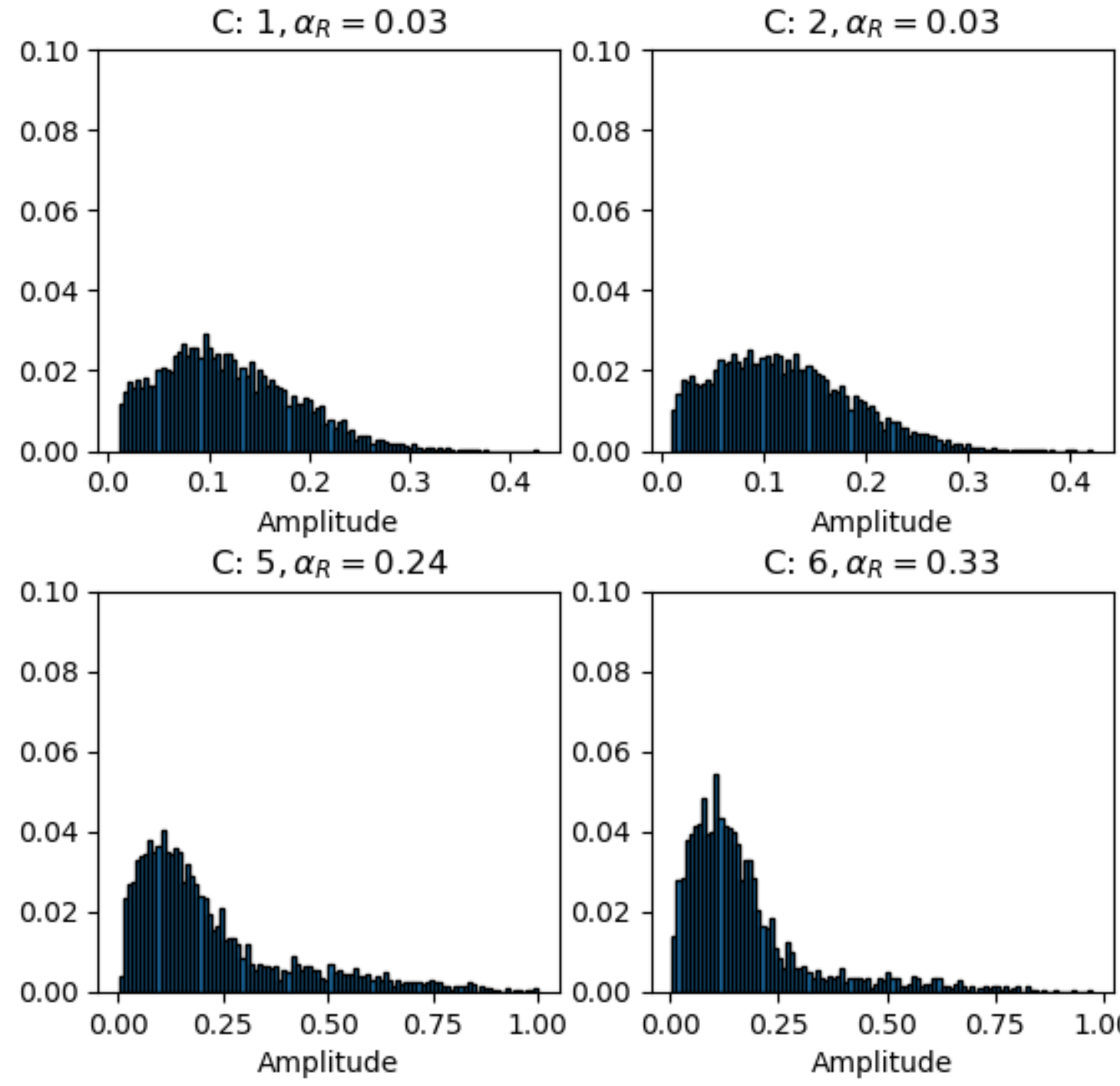}
	\caption{Noisy data, Gaussian $(\mu = 0.1A_{max}, \sigma = 0.1A_{max})$}
	\label{fig_stat2}
\end{subfigure}	
	\caption{Frequency distribution of amplitudes within 4 out of 8 partitions for (a) the noisefree image with 5 events shown in Fig. \ref{fig_syn}, and (b) its noisy variant with added Gaussian noise. For the partitions that contain an event, $\alpha_R$ is relatively higher.}
	\label{fig_stats}
\end{figure*}

After the data domain has been divided into a number of segments, there exist certain segments having cluster-centroids (approximately) aligned with the event-centers, and other segments with no events. As a next step, the segments which do not encompass an event, need to be discarded. For this purpose, the statistical distribution of amplitudes is studied within every segment, and based on a certain mathematical criterion defined by $\alpha_R$, the respective segment is accepted or discarded. Fig. \ref{fig_stat1} shows the distribution of amplitudes for the various segments shown in \mbox{Fig. \ref{fig_case1_kmeans}}. Here, it is straightforward to identify the partitions that correspond to an event (Clusters 4 and 7) and the partitions which are empty (3 and 8). However, for noisy data, isolating the empty clusters is not easy (Fig. \ref{fig_stat2}). More details associated with the elimination criterion are discussed in \mbox{Section \ref{sec_method}}.

\begin{figure}[t]
	\centering
	\begin{subfigure}{0.4\textwidth}
		\centering
		\includegraphics[height=4.8cm, width=5.3cm]{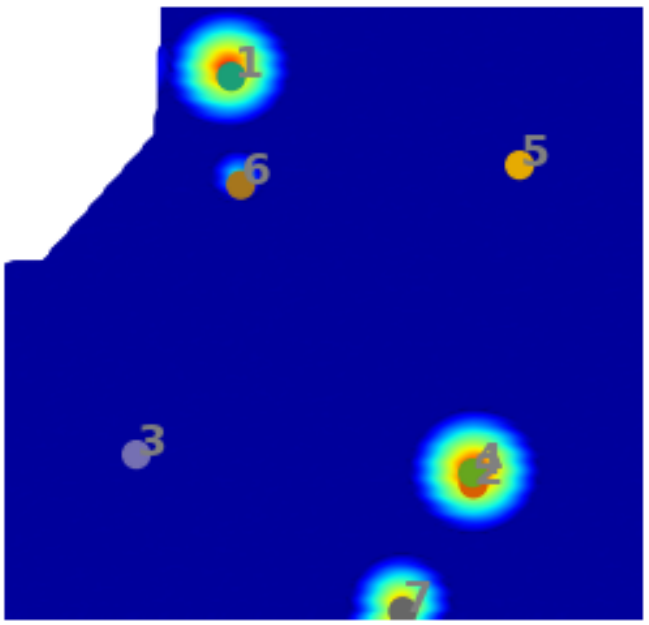}
		\caption{Identified events}
	\end{subfigure}
	\begin{subfigure}{0.4\textwidth}
		\centering
		\includegraphics[height=4.8cm, width=5.3cm]{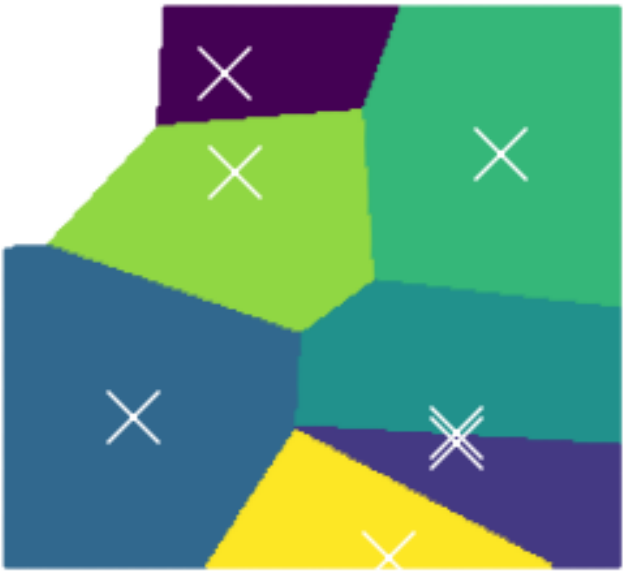}
		\caption{Optimized clusters}
	\end{subfigure}			
	\caption{Results obtained from modified $k$-means clustering during Cycle 2 of ICS, showing (a) the remaining 4 synthetic events and the optimized locations of 7 cluster centers, and (b) the 7 segments with their centers marked using `$\times$'.}
	\label{fig_case1_kmeans2}
\end{figure}

Next, the remaining segments are studied to identify whether there are partitions that correspond to the same event. Fig. \ref{fig_kmeans_elim} shows the segments that are left after elimination of empty clusters in Fig. \ref{fig_stat1}. It can be seen that there are two clusters which are formed around the same event, and these need to be merged. This is done by analyzing the `span' of each event, where span refers to the space in which the disturbance in amplitude has been created by the event. If the center of one event lies within the span of another one, both the events are merged as a single event. Thus, in \mbox{Fig. \ref{fig_kmeans_elim}}, clusters 2 and 7 get merged.

The process from clustering to merging the common partitions constitutes one cycle of ICS process. Among the identified events, the strongest event is selected, and the data corresponding to its segment is deleted from the data domain. For the events identified in Fig. \ref{fig_kmeans_elim}, Segment 1 is identified to be containing the strongest event. Thus, the data for this segment is removed from the domain, and the second cycle of ICS is executed to identify the next event. \mbox{Fig. \ref{fig_case1_kmeans2}} shows the identified events and segments obtained from clustering during Cycle 2 of ICS. In this manner, the ICS process is repeated over several cycles, till no more event can be identified. From Cycle 2, every new event selected at the completion of the ICS process, is checked with the already selected events for possible overlap.
 
\subsection{Method} 
\label{sec_method}
The process of iteratively identifying the events can be broken down into 4 steps, and these have briefly been discussed in Section \ref{sec_desc}. Below, we discuss the details related to each of these steps.

\textbf{Modified $k$-means clustering.} During every cycle of ICS methodology, the domain is segmented using a clustering approach. Since only circular (spherical) events are studied in this paper, a modified form of $k$-means clustering has been implemented. The choice of $k$ here depends on a priori knowledge of the problem that we are dealing with. However, we have observed that the algorithm itself is not very sensitive to the choice of $k$. As a rule of thumb, we choose $k$ to be approximately 2-3 times the maximum number of events that can generally be expected to occur in the domain. This can be further lower for data where the amplitude of background noise is significantly lower than that of the event. For example, for the synthetic test cases shown in Fig. \ref{fig_eg_kmeanspp}, even $k = 8$ is found to be sufficient. A brief discussion on the initial choice of $k$ is presented in \mbox{Section \ref{sec_discuss}}. 

For the modified $k$-means clustering, the locations of the initial centroids are chosen using the $k$-means++ \mbox{approach \cite{Arthur2007}}. Let the amplitude distribution be represented by a set of $N$ data points $\mathcal{X} \subset \mathbb{R}^d$, where $d$ refers to dimensions of the data. In the $k$-means++ approach, the cluster centroid $c_1$ is chosen uniformly at random from $\mathcal{X}$. The centroids $c_2, c_3, \hdots, c_k$ are then iteratively chosen, and for every next centroid $c_i$, any point $\mathbf{x}' \in \mathcal{X}$ can be chosen with probability $\frac{D(\mathbf{x}')^2}{\sum_{\mathbf{x} \in \mathcal{X}} D(\mathbf{x})^2}$. Here, $D(\mathbf{x})$ denotes the distance of data point $\mathbf{x}$ to the nearest amongst cluster centroids $c_1, c_2, \hdots, c_{i-1}$. The process is continued until all the $k$ cluster-centroids have been chosen.

Based on $k$-means++, $\mathcal{X}$ gets partitioned into $k$ segments denoted as $\mathcal{S} = {S_1, S_2, \hdots, S_k}$. Next, a variant of the traditional $k$-means clustering method is applied to optimize the cluster-centroids \mbox{$\mathcal{C} = \{c_1, c_2, \hdots, c_k\}$}. This involves solving a zeroth order optimization problem which is stated as 
\begin{equation}
	\underset{\mathcal{C}}{\min} \enskip \mathcal{R}_N(\mathcal{C}) =  \sum_{j = 1}^{N} \underset{1\leq i\leq k}{\min} \enskip \phi(A(\mathbf{x})) \lVert \mathbf{x}_j - c_i \rVert^2,
	\label{eq_obj}
\end{equation}
where $\phi(\cdot)$ denotes an amplitude functional and $A(\mathbf{x})$ denotes amplitude-value of data point $\mathbf{x}$. Each iteration of the clustering method involves two updates which are as follows:
\begin{enumerate}
	\item \emph{Assignment step}: Each data point $\mathbf{x} \in \mathcal{X}$ needs to be mapped to one of the clusters in $\mathcal{C}$ to obtain a new set of partitions $\mathcal{S}$. Mathematically, data point $\mathbf{x}_j$ is mapped to partition $S_i$, where $i = \underset{e \in [1, k]}{\arg\min} \phi(A(\mathbf{x}))\lVert \mathbf{x} - c_e \rVert^2$.
	\item \emph{Update step}: After each data point has been assigned to one of the partitions in $\mathcal{S}$, the locations of centroids in $\mathcal{C}$ are updated as
		\begin{equation}
			c_i = \frac{\sum_{\mathbf{x} \in S_i} \phi(A(\mathbf{x})) \lVert \mathbf{x} - c_i \rVert^2\mathbf{x}}{\sum_{\mathbf{x} \in S_i} \phi(A(\mathbf{x})) \lVert \mathbf{x} - c_i \rVert^2}.
		\end{equation}   
\end{enumerate}
These two steps are repeated until the value of  objective $\mathcal{R}_N(\mathcal{C})$ goes below a certain minimum threshold, or some other stopping criterion is reached. Note that unlike the initial assignment done using $k$-means++, the amplitude of data points has been taken into account during the assignment as well as update steps of ICS.
\begin{figure}[t]
	\centering
	\begin{subfigure}{0.3\textwidth}
		\centering
		\includegraphics[height=4cm, width=4cm]{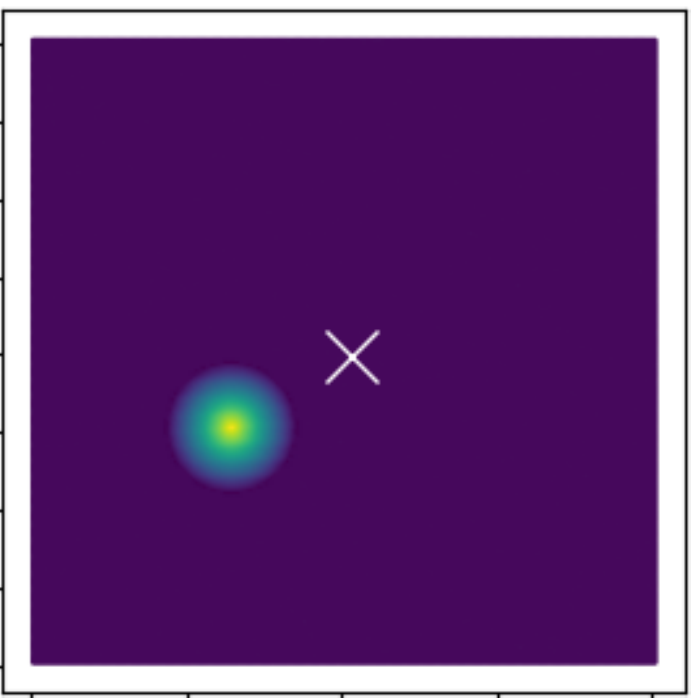}
		\caption{$p = 0$}
	\end{subfigure}
	\begin{subfigure}{0.3\textwidth}
		\centering
		\includegraphics[height=4cm, width=4cm]{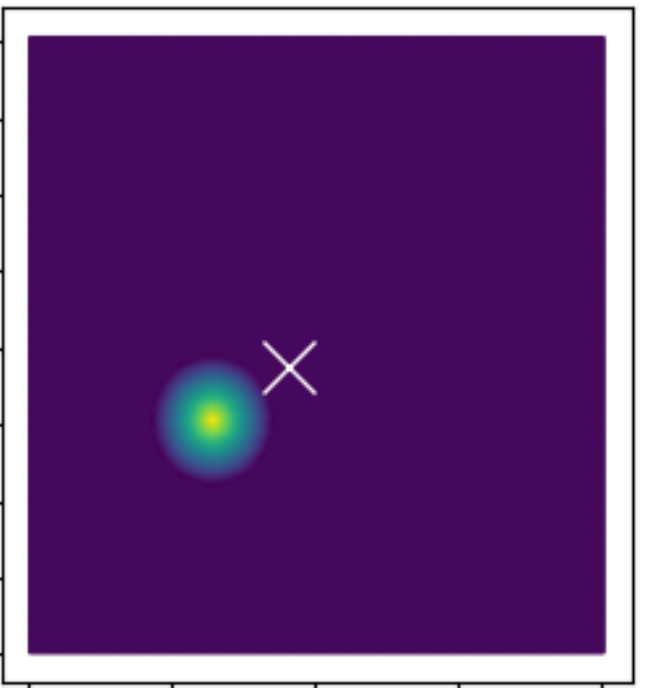}
		\caption{$p =3$}
	\end{subfigure}	
	\begin{subfigure}{0.3\textwidth}
		\centering
		\includegraphics[height=4cm, width=4cm]{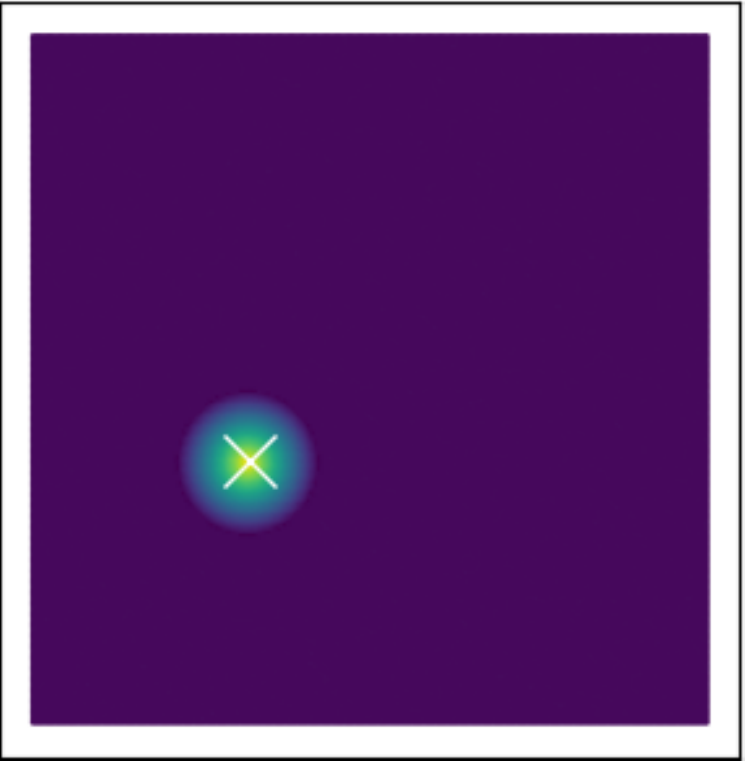}
		\caption{$p = 5$}
	\end{subfigure}			
	\caption{Synthetic 2D images, each containing one event, showing the location of cluster-centroids obtained for the 3 cases using modified $k$-means clustering \mbox{($\phi(A(\mathbf{x})) = A(\mathbf{x})^p$)} with different values of $p$.}
	\label{fig_amppow}
\end{figure}

A better understanding on the role of $\phi(A(\mathbf{x}))$ can be obtained from Fig. \ref{fig_amppow}. For simplicity let us assume that $\phi(A(\mathbf{x})) = A(\mathbf{x})^p$, where $p$ is a power term used to adjust the weight on amplitude values. For this example, \mbox{$A(\mathbf{x}) > 1 \enskip \forall \enskip \mathbf{x} \in \mathcal{X}$}. In Fig. \ref{fig_amppow}, we have one event in a 2D image, and the goal of ICS approach is to identify the event-center. The data points are assumed to be uniformly distributed in the domain. With $p = 0$, the clustering approach is independent of amplitude-values, due to which the cluster centroid is identified at approximately the geometric center of the image. With $p=1$, amplitudes are provided certain weight, due to which the cluster centroid shifts towards the event. Adding more weight with $p = 5$, the cluster centroid overlaps with the event. Thus, when $p$ is chosen properly, the modified clustering approach can detect the event.

However, the formulation for $\phi$ mentioned above will be sensitive to the values of $A$, and we formulate it differently for our problems to make it more robust. The amplitudes are normalized between 0 and 1, and $\phi(A(\mathbf{x})) = e^{pA(\mathbf{x})}$, where $p$ is a exponential scaling term. For the examples shown in this paper, $p$ is set to 16 based on observations from a set of numerical experiments.

\textbf{Elimination of empty clusters.} After the data has been divided into partitions $\mathcal{S}$, the empty partitions need to be separated from those that contain the events. As stated earlier in Section \ref{sec_desc}, this is achieved by analyzing the statistical distribution of amplitude-values inside every partition, and comparing them against a certain threshold. Unlike the traditional approach of choosing a certain value of $A(\mathbf{\mathbf{x}})$ for thresholding, functional $\phi(A(\mathbf{x}))$ is analyzed. 
\begin{figure}[t]
	\centering
	\includegraphics[scale=0.9]{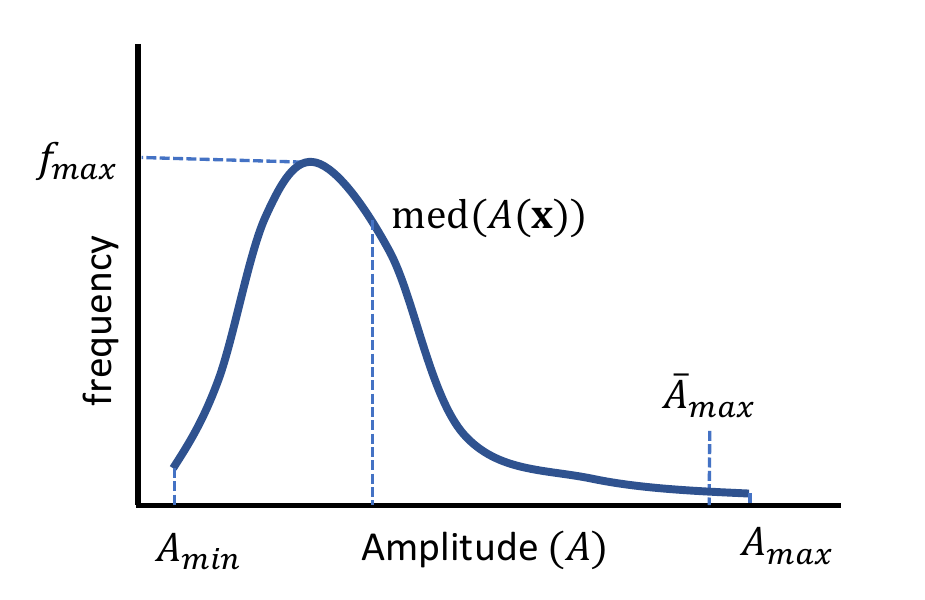}
	\caption{Schematic representation for the distribution of amplitude-values inside a partition obtained from clustering. The symbols $f_{max}$, $\text{med}(\cdot)$ and $\bar{A}_{max}$ refer to highest frequency value for a partition, median functional, and the maximum value of amplitude obtained after the elimination of 0.1\% of the amplitudes from the right-tail of the distribution. For an event defined using Gaussian distribution or a linear decay from the center, the distribution of amplitudes resembles the distribution shown here.}
	\label{fig_elim1}
\end{figure}

Fig. \ref{fig_elim1} shows a schematic histogram plot for the amplitude values observed in a certain partition $S_i$. Ideally, for an event characterized by a Gaussian distribution of amplitudes or a linear decay away from the event-center, the distribution of amplitudes is expected to resemble a right skewed distribution with very low values of frequency at the right tail. To decide whether the distribution shown in Fig. \ref{fig_elim1} corresponds to an event or not, the rejection factor $\alpha_R$ needs to be defined. Mathematically, this term is defined as 
\begin{equation}
	\alpha_R^{(i)} = \frac{f_{max}}{\text{med}(A(\mathbf{x}))} \bar{A}(\mathbf{x})^2 \enskip \forall \enskip \mathbf{x} \in S_i,
	\label{eq_alpha_r}
\end{equation}
where, superscript $i$ is an index term used to refer to partition $S_i$, $f_{max}$ denotes the maximum value of frequency for the distribution of amplitudes discretized into 100 bins, and $\text{med}(\cdot)$ denotes the median functional. Further, $\bar{A}(\mathbf{x})$  calculates the maximum value of amplitude in $S_i$ by setting statistical significance level at 0.999, which means 0.1\% of the values at the right tail of the distribution in Fig. \ref{fig_elim1} are ignored. 

To understand the motivation for choosing the expression in Eq. \ref{eq_alpha_r}, we refer back to Fig. \ref{fig_elim1}. Here, the maximum value of amplitude ($A_{max}$) plays role in deciding whether the distribution corresponds to an event or not (as in \mbox{Fig. \ref{fig_stat1})}. For partitions which do not contain an event, $A_{max}$ is expected to be equal to the maximum noise-level in the data, and for partitions with event, it corresponds to that of the event. There is a possibility that random noise-spikes are contained in the data whose amplitudes are higher than the event amplitude itself. To avoid capturing these spikes as event, 0.1\% of the data on the right tail side of the distribution is categorized as statistically insignificant, and $A_{max}$ is replaced by $\bar{A}_{max}$. Here, $\bar{A}_{max}$ characterizes the strength of the event, and ranks it against other events in the iterative selection criterion. In this sense, it can be considered as a \emph{global criterion} for event detection.

For weak events, $\bar{A}_{max}$ is low, which reduces $\alpha_R$ for that partition. For such partitions, $\frac{f_{max}}{\text{med}(A(\mathbf{x}))}$ helps to differentiate an event from background noise.  Unlike an actual event, the amplitudes in background noise are uniformly distributed over a range of values, due to which its $f_{max}$ is expected to be lower than that of an event. A better understanding of this can be obtained from Fig. \ref{fig_stat2}. This figure shows the histogram plots for partitions obtained using ICS for one of the examples shown in \mbox{Fig. \ref{fig_eg_kmeanspp}}, but with added Gaussian noise ($\mu = 0.1A_{max}, \sigma = 0.1A_{max}$). Among the 4 clusters shown in Fig. \ref{fig_stat2}, Clusters 5 and 6 correspond to partitions containing events, thus clearly have higher values of $f_{max}$. Also, for partitions in $\mathcal{S}$ that contain an event (\emph{e.g.} Clusters 5 and 6), median of the distribution will be shifted more towards left compared to that in partitions that contain only background noise. In this manner, $\frac{f_{max}}{\text{med}(A(\mathbf{x}))}$ together with  $\bar{A}_{max}$ defines $\alpha_R$ which is further used to eliminate the empty partitions.   	

Partition $S_i$ is then rejected if $\alpha_R^{(i)} < \alpha_R^{(c)}$, where $\alpha_R^{(c)}$ refers to a cut-off threshold. The value of   $\alpha_R^{(c)}$ is chosen  based on numerical experiments, and in this paper, it is set to 0.1. Further discussion on the sensitivity of this parameter as well as on how to choose it, is presented in Section \ref{sec_discuss}. With the partitions left after eliminating the empty ones, the next step is to merge the overlapping partitions. 

\begin{figure}[t]
	\centering
	\includegraphics[scale=0.9]{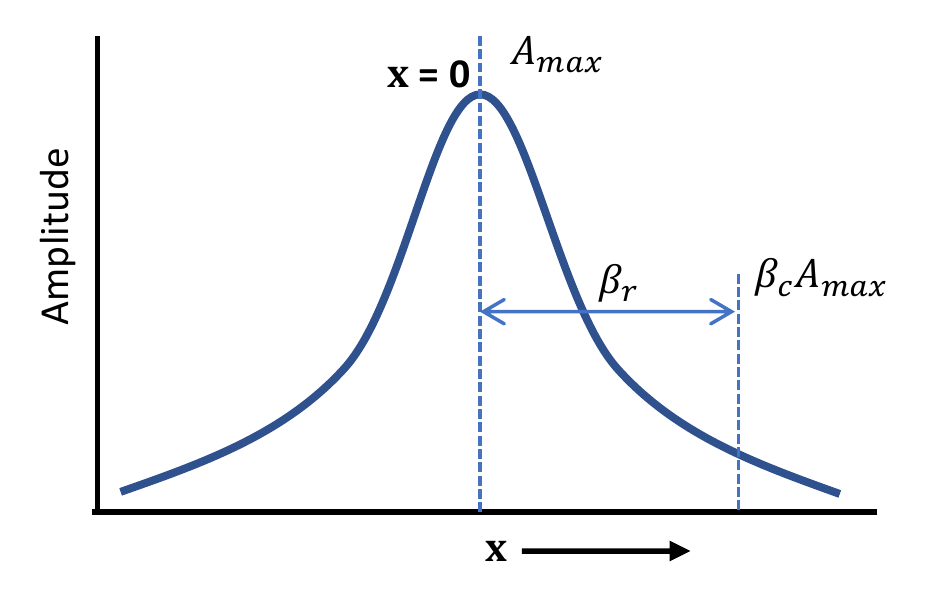}
	\caption{Schematic representation for a one-dimensional distribution of amplitude values. Here, span radius $\beta_r$ denotes the distance from the center of the event to the point where amplitude reduces to $\beta_r$ times $A_{max}$. The span radius defines the extent of significance for the event, and any other event-center that lies within the span is constituted as a part of this event.}
	\label{fig_merge1}
\end{figure}

\textbf{Merge overlapping clusters.} Partitions which share the same event need to be identified. For this purpose, the distribution of amplitudes in space (for 2D and 3D) and time (for 4D) needs to be studied and the extent of overlap has to be identified. Fig. \ref{fig_merge1} shows the schematic representation of amplitude distribution for 1D space, and an approximate Gaussian distribution is assumed. Next, we define the term span radius $\beta_r$, which denotes the distance from the center of an event up to which the event is considered to span. Here, $\beta_c \in (0, 1]$ is a fractional term used to control the span. For example, to assume that the event is significant up to the point where $A(\mathbf{x})$ decays to 40\% of its maximum, $\beta_c = 0.4$. Thus, in simple terms, $\beta_r$ can be defined as the distance from the center of an event, where $A$ decays to $\beta_c$ fraction of $A_{max}$.

Although from Fig. \ref{fig_merge1}, it seems very straightforward to compute $\beta_r$, the distributions are not perfectly Gaussian and frequently contain noise. Thus, there will be multiple values of $\beta_r$ that satisfy the requirement. Moreover, for robustness, we need to search for values within the range $(\beta_c A_{max} - \epsilon, \beta_c A_{max} +\epsilon)$ rather than $\beta_c A_{max}$. Here, $\epsilon$ is a parameter that defines an amplitude interval for calculating the span, and in this study, it has been set to 0.01. More discussions on choosing $\beta_c$ and $\epsilon$ follow in Section \ref{sec_discuss}. Thus, for the $i^{\text{th}}$ partition, $\beta_r^{(i)}$ is defined as
\begin{align}
	\beta_r^{(i)} = \lVert \beta_r^{(t)} \rVert_2 \enskip \forall \enskip \{t: t \in [1, N], \mathbf{x}^{(t)} \in \mathcal{X},\nonumber \\ 
 \beta_c A_{max} - \epsilon \leq A(\mathbf{x}^{(t)}) \leq \beta_c A_{max} + \epsilon\}.
\end{align}

Once $\beta_r$ has been calculated for all the partitions, partition $p$ is merged to partition $q$, if:
\begin{equation}
	\lVert c_p - c_q \rVert_2 \leq \beta_r^{(q)}, \enskip \forall \enskip p, q \in [1, k], p \neq q, \bar{A}_{max}^{(p)} < \bar{A}_{max}^{(q)}.  
\end{equation}
Accordingly, $S_q = S_q \cup S_p$ and $S_p = \phi$. Further, $\beta_r^{(q)}$ is updated based on the new set of data points in $S_q$.

\textbf{Select the strongest event.} After the overlapping partitions have been merged, each of the partitions $S_i \in \mathcal{S}$ is expected to contain an event. As defined earlier, our ICS methodology selects only one event per cycle. To understand the motivation for doing so, we look at Eq. \ref{eq_obj}. Since the objective functional here depends on $A$ as well, there would be a preference to form clusters around the stronger events. In this scenario, if there is an outlier event which is far stronger than the rest of the events, most of the partitions will be clustered around it, and the weaker events of the data will be shadowed.  

The iterative scheme in ICS selects the strongest event during the first cycle, and the corresponding partition is then removed from $\mathcal{S}$ for the next ICS cycle. This facilitates improved clustering around the weaker events during the later cycles. Eventually, most of the events can be identified over a series of cycles. Let $S_l$ denote the partition selected during the $l^{\text{th}}$ cycle. Assuming that the stopping criterion has been reached after $m$ cycles of ICS, the centroids $c_l \enskip \forall \enskip l \in [1, m], l \in \mathbb{Z}^+$ corresponding to partitions $S_l$ correspond to the identified events.    

\section{Applications}
\label{sec_appl}

\begin{figure*}
	\centering
	\hspace{-2em}
	\begin{subfigure}{0.3\textwidth}
		\centering
		\includegraphics[height=4.8cm, width=5.3cm]{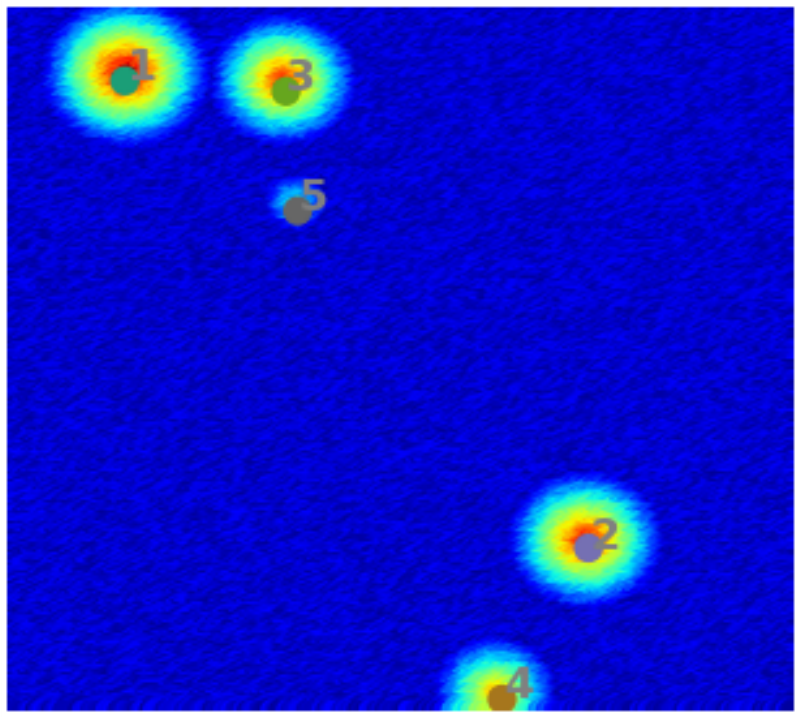}
		\caption{Random}
		\label{fig_rand1}
	\end{subfigure} \enskip
	\begin{subfigure}{0.3\textwidth}
		\centering
		\includegraphics[height=4.8cm, width=5.3cm]{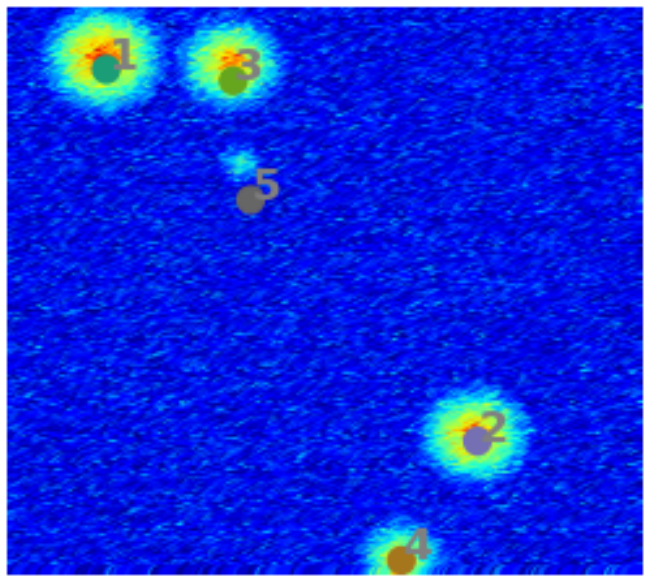}
		\caption{Normal}
		\label{fig_norm1}
	\end{subfigure}	\enskip	
	\centering
	\begin{subfigure}{0.3\textwidth}
		\centering
		\includegraphics[height=4.8cm, width=5.3cm]{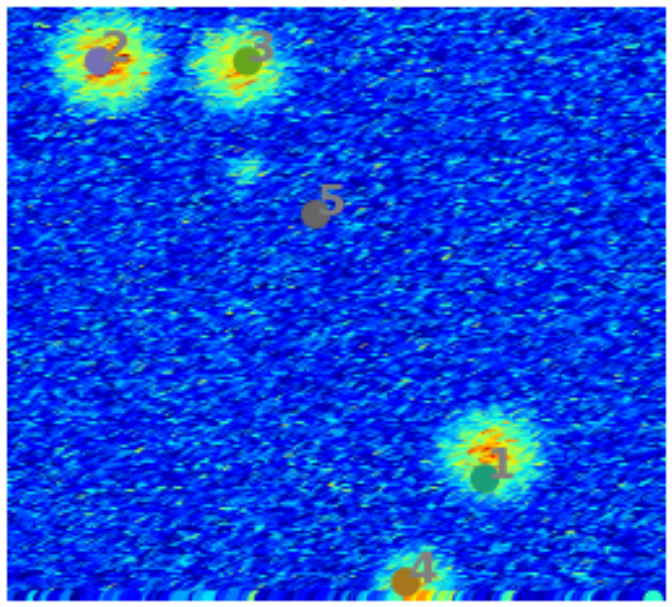}
		\caption{Normal}
		\label{fig_rand2}
	\end{subfigure}\enskip
	\begin{subfigure}{0.3\textwidth}
		\centering
		\includegraphics[height=4.8cm, width=5.3cm]{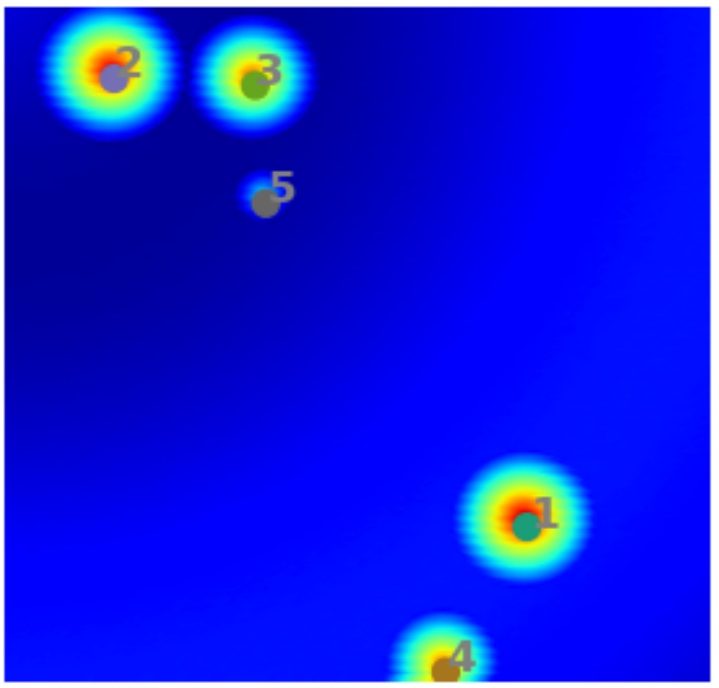}
		\caption{Periodic}
		\label{fig_per1}
	\end{subfigure}	\enskip
	\begin{subfigure}{0.3\textwidth}
		\centering
		\includegraphics[height=4.8cm, width=5.3cm]{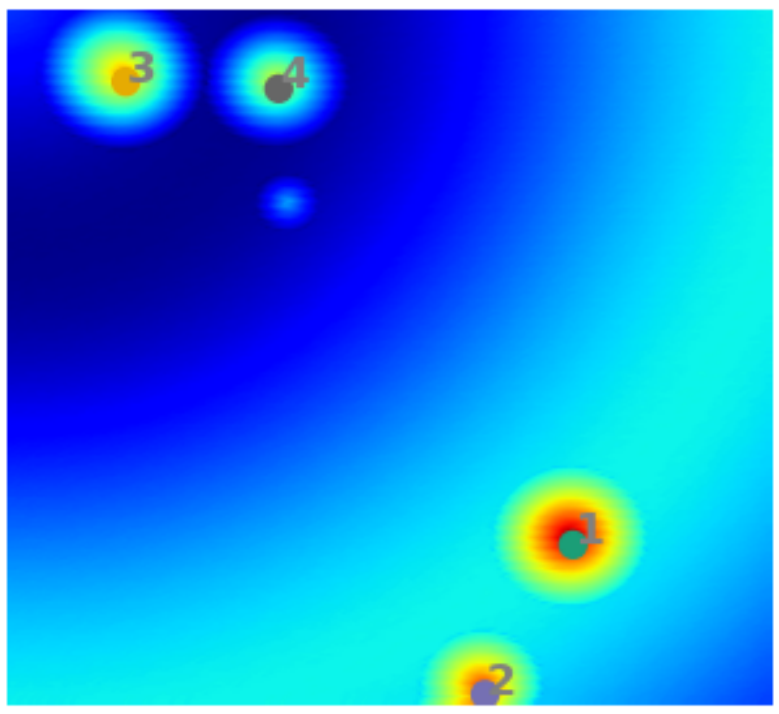}
		\caption{Periodic}
		\label{fig_per2}
	\end{subfigure}	
	\caption{Synthetic 2D images showing 5 events each and locations of the events identified using ICS method. The types of noise are: (a) uniform random noise ($A_{max}^{noise} = 0.1 A_{max}$), (b) uniform normal noise ($\mu = 0.1A_{max}, \sigma = 0.1A_{max}$), (c) uniform normal noise ($\mu = 0.1A_{max}, \sigma = 0.2A_{max}$), (d) sinusoidal noise ($\lambda = (L_x^2 + L_y^2)^{1/2}$ and $A_{max}^{noise} = 0.25 A_{max}$), and (e) sinusoidal noise ($\lambda = (L_x^2 + L_y^2)^{1/2}$ and $A_{max}^{noise} = A_{max}$). In (b) and (c), although the centroid of Cluster 5 does not overlap with the nearby event, both of them fall within the same partition.}
	\label{fig_syn5}
\end{figure*}

\begin{figure}[t]
	\centering
	\begin{subfigure}{0.4\textwidth}
		\centering
		\includegraphics[height=6cm, width=6.5cm]{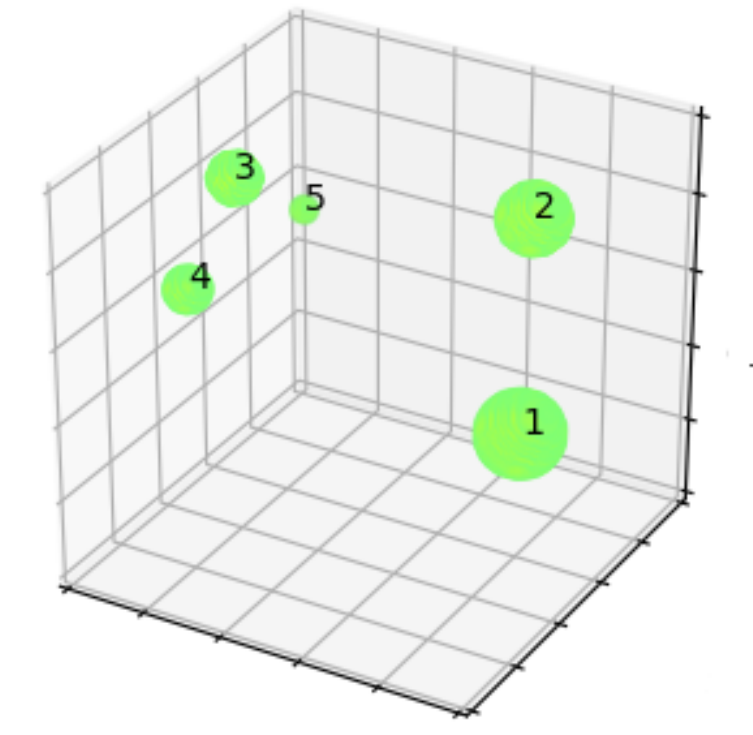}
		\caption{Synthetic (5 events)}
		\label{fig_3d_1}
	\end{subfigure}
	\begin{subfigure}{0.4\textwidth}
		\centering
		\includegraphics[height=6cm, width=6.5cm]{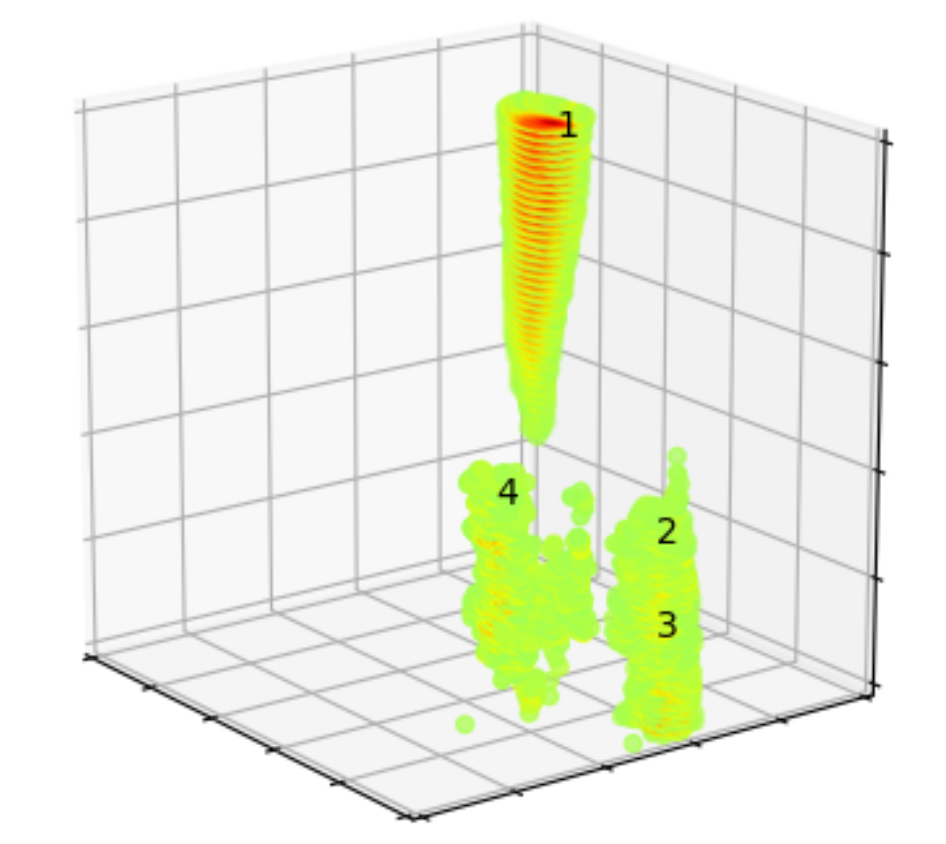}
		\caption{Real (3+ events)}
		\label{fig_3d_2}
	\end{subfigure}
	\caption{Examples of 3D images with events and the event-centroids identified using ICS method. For the real dataset, 3 events had previously been marked manually. For visualization purpose, only amplitudes above 0.5$A_{max}$ have been shown.}
	\label{fig_3d}
\end{figure}

To demonstrate the applicability of ICS, we consider examples of 2D and 3D images. For 2D cases, a total of 140 synthetic images (noisefree as well as containing certain noise) have been considered, and average accuracies are reported in Fig. \ref{fig_nk_stat} for several values of $N_c$. An example application of ICS on 2D noisefree images has been demonstrated in Fig. \ref{fig_case1_kmeans}. 

Fig. \ref{fig_syn5} further demonstrates the potential of ICS under different noise levels for uniform random, uniform normal and periodic noises. Details related to types and levels of the added noise are described in the caption of Fig. \ref{fig_syn5}. For all the cases, it is observed that ICS can resolve all the major events observed in the images. However, for high levels of noise (as in Figs \ref{fig_syn5}e), one of the weak events cannot be identified. For rest of the cases, we observe that a correct number of events have been identified. At low noise-levels, the identified event locations are correct. However, for the cases of Gaussian noise considered here (Figs. \ref{fig_syn5} c and d), the identified locations of Event 5 seem to deviate from the actual value. This happens primarily because for any partition, we assume the point with highest amplitude value to be the event centroid. However, in the presence of noise, this assumption might not be true, and a direction of improvement would be to correctly identify the event location within the active partition.  \\
Further, two 3D test cases have been considered as shown in Fig. \ref{fig_3d}, and here, only amplitude values above $0.5 A_{max}$ have been displayed. The synthetic image shown in Fig. \ref{fig_3d_1} consists of 5 events with no added noise. Using ICS method, all the 5 events could been identified successfully. Fig. \ref{fig_3d_2} shows the application of ICS method on a real dataset of earthquake-events. Based on manual interpretation, 3 events had been identified in the past for this case. Using ICS, we identified 4 events, and further manual analysis confirmed that the additional event identified by ICS is indeed a real event. Clearly, in 3D or higher, human interpretations are limited due to visual constraints, and this numerical example demonstrates that such a limitation can be overcome using ICS.    
\section{Discussions}
\label{sec_discuss}

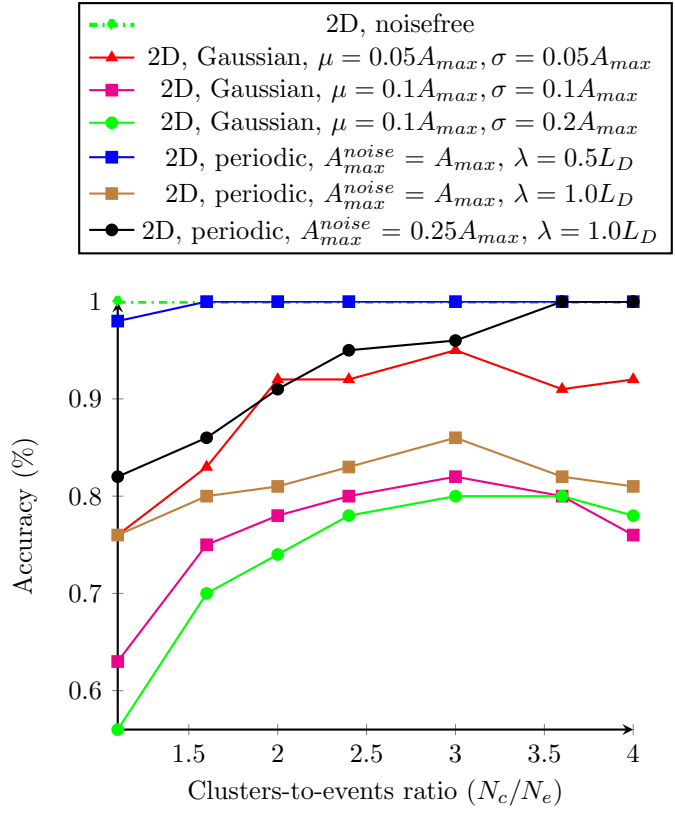
\begin{figure} 
	\begin{center}
		\begin{tikzpicture}[scale = 1]
		\begin{axis}[%
		ymax = 1,
		thick, 
		axis x line=bottom,
		axis y line=left,
		legend style={at={(0.5,1.7)},anchor=north},
		xlabel = {Clusters-to-events ratio ($N_c/N_e$)},
		ylabel = {Accuracy (\%)}
		]
		\addplot[mark=triangle*, mark options={scale=1}, dashdotted, green, ultra thick] table[x=case2_nc, y = accuracy2] {data/cluster_choice.txt};
		\addplot[mark=triangle*, mark options={scale=1}, solid, red,  thick] table[x=case3_nc,y = accuracy3] {data/cluster_choice.txt};		
		\addplot[mark=square*, mark options={scale=1}, magenta, thick] table[x=case1_nc,y = accuracy1] {data/cluster_choice.txt};
		\addplot[mark=*, mark options={scale=1}, solid, green, thick] table[x=case4_nc,y = accuracy4] {data/cluster_choice.txt};
		\addplot[mark=square*, mark options={scale=1}, solid, blue, thick] table[x=case5_nc,y = accuracy5] {data/cluster_choice.txt};
		\addplot[mark=square*, mark options={scale=1}, solid, brown, thick] table[x=case6_nc,y = accuracy6] {data/cluster_choice.txt};
		\addplot[mark=*, mark options={scale=1}, solid, black, thick] table[x=case7_nc,y = accuracy7] {data/cluster_choice.txt};
		\addlegendentry{2D, noisefree}
		\addlegendentry{2D, Gaussian, $\mu=0.05A_{max}, \sigma=0.05A_{max}$}
		\addlegendentry{2D, Gaussian, $\mu=0.1A_{max}, \sigma=0.1A_{max}$}
		\addlegendentry{2D, Gaussian, $\mu=0.1A_{max}, \sigma=0.2A_{max}$}
		\addlegendentry{2D, periodic, $A_{max}^{noise}$ = $A_{max}$, $\lambda = 0.5L_D$}
		\addlegendentry{2D, periodic, $A_{max}^{noise}$ = $A_{max}$, $\lambda = 1.0L_D$}
		\addlegendentry{2D, periodic, $A_{max}^{noise}$ = 0.25$A_{max}$, $\lambda = 1.0L_D$}
		\end{axis}
		\end{tikzpicture}
	\end{center}
	\caption{Accuracy versus clusters-to-events ratio ($N_c/N_e$) for several synthetic test cases. Here, $N_c$ denotes the number of clusters chosen for the first cycle of ICS and $N_e$ refers to the number of events in the data. For the datasets containing periodic noise, sinusoidal noise has been added to the data, and $\lambda$ and $L_D$ denote the wavelength used and length of the major diagonal for the data domain.}
	\label{fig_nk_stat}
\end{figure}

There are a few parameters that control the performance of the proposed ICS methodology. These include initial number of clusters $N_c$, amplitude power $p$, rejection factor $\alpha_R^{(c)}$, decay factor $\beta_c$ and $\epsilon$. These parameters have been chosen statistically, and their optimal values have been chosen through a number of simulations on 2D datasets. Due to constraint on the length of this paper, we discuss only one of these parameters. 

Fig. \ref{fig_nk_stat} shows the affect of $N_c/N_e$, where $N_e$ refers to number of events in the simulated data. A total of 20 test cases of 2D synthetic images were used, each containing between 3 and 6 events of varying strength. Each of these images was duplicated 6 times with different noise types and levels to generate 120 noisy cases. Further, for every case, 10 simulations have been run, and the average ICS accuracies are reported in Fig. \mbox{\ref{fig_nk_stat}}. A general observation is that sufficiently good accuracies are obtained for $N_c/N_e$ values in between 2 and 3. For higher values, the domain is analyzed using too many partitions, which leads to insufficient information being contained in every partition, and this leads to loss in accuracy.  

Similar tests have been carried out for other parameters, and here we only report the optimal choices observed. For amplitude power $p$, a value between 13 and 17 works well. Similarly for $\alpha_R^{(c)}$, $\beta_c$ and  $\epsilon$, the recommended values are 0.05-0.15, 0.3-0.5, 0.1-0.2, respectively. While the other two parameters are not so important, $\alpha_R^{(c)}$ is very critical since it plays an important role in deciding whether a partition contains an event or not. To some extent, its value depends on the noise level in the data, and in future we aim to look into this aspect.

\section{Conclusion and Future work }
In this paper, we presented ICS, an iterative clustering based segmentation approach for the detection of events in image data. Over a sequence of cycles, ICS partitions an image and analyzes whether each partition contains an event or not. Aiming at automated real time detection of events in continuously streaming data, the applicability of ICS has been demonstrated on 2D and 3D datasets. \\
In the future work, we aim at exploring the application for higher dimensions. Further, we look forward to using ICS on microseismic data for the automated real time detection of earthquakes. 

\bibliography{egbib}
\bibliographystyle{plain}
\end{document}